\pdfoutput=1

\documentclass[11pt]{article}

\usepackage[]{acl}

\usepackage{times}
\usepackage{latexsym}
\usepackage{nicefrac}
\usepackage{algorithm}
\usepackage[noend]{algpseudocode}
\usepackage{algpseudocode}
\algrenewcommand\algorithmicrequire{\textbf{Input:}}
\algrenewcommand\algorithmicensure{\textbf{Output:}}
\usepackage{mathtools}
\usepackage{xspace}
\usepackage{siunitx}
\usepackage{enumitem}
\usepackage{amsfonts,amsmath,amssymb,bm,xifthen}
\usepackage{caption}
\usepackage{subcaption}
\usepackage{multirow}

\usepackage[T1]{fontenc}

\usepackage[utf8]{inputenc}

\usepackage{microtype}

\usepackage{inconsolata}
\usepackage{multicol}

\title{Distantly-Supervised Joint Extraction with Noise-Robust Learning}

\author{
Yufei Li$^{1}$
\xspace\xspace    
Xiao Yu$^{2}$
\xspace\xspace
Yanghong Guo$^{3}$
\xspace\xspace
Yanchi Liu$^{4}$
\xspace\xspace    
Haifeng Chen$^{4}$
\xspace\xspace    
Cong Liu$^{1}$\\
$^{1}$UC Riverside\xspace\xspace
$^{2}$Stellar Cyber\xspace\xspace
$^{3}$UT Dallas\xspace\xspace
$^{4}$NEC Labs America\\
$^{1}$\texttt{\{yli927,congl\}@ucr.edu}\xspace\xspace  $^{2}$\texttt{xyu@stellarcyber.ai}\xspace\xspace
$^{3}$\texttt{yxg190031@utdallas.edu}
\\
$^{4}$\texttt{\{yanchi,haifeng\}@nec-labs.com}
}

\begin{document}
\maketitle
\begin{abstract}
Joint entity and relation extraction is a process that identifies entity pairs and their relations using a single model. 
We focus on the problem of joint extraction in distantly-labeled data, whose labels are generated by aligning entity mentions with the corresponding entity and relation tags using a knowledge base~(KB). 
One key challenge is the presence of noisy labels arising from both incorrect entity and relation annotations, which significantly impairs the quality of supervised learning. 
Existing approaches, either considering only one source of noise or making decisions using external knowledge, cannot well-utilize significant information in the training data.
We propose DENRL, a generalizable framework that 1)~incorporates a lightweight transformer backbone into a sequence labeling scheme for joint tagging, and 2)~employs a noise-robust framework that regularizes the tagging model with significant relation patterns and entity-relation dependencies, then iteratively self-adapts to instances with less noise from both sources. 
Surprisingly, experiments\footnote{Our code is available at \url{https://github.com/yul091/DENRL}.} on two benchmark datasets show that DENRL, using merely its own parametric distribution and simple data-driven heuristics, outperforms large language model-based baselines by a large margin with better interpretability. 
\end{abstract}

\section{Introduction}

\begin{figure}
    \centering
    \includegraphics[width=0.49\textwidth]{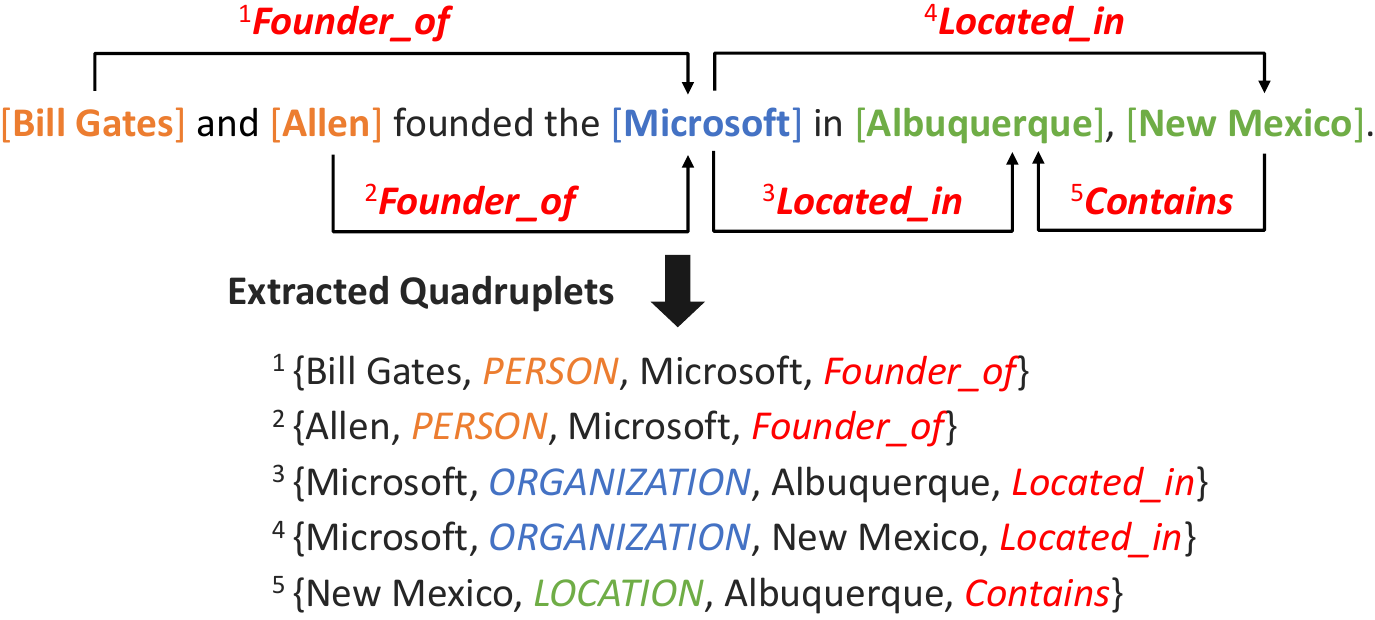}
    \caption{An example of joint extraction on a sentence with multiple relations that share the same entity, e.g., ``\emph{Microsoft}'' in both the third and the forth relations.}
    \label{fig:example}
    \vspace{-0.5cm}
\end{figure}

Joint extraction aims to detect entities along with their relations using a single model (see Figure~\ref{fig:example}), which is a critical step in automatic knowledge base construction~\cite{yu2019joint}. 
In order to cheaply acquire a large amount of labeled joint training data, distant supervision~(DS)~\cite{mintz2009distant} was proposed to automatically generate training data by aligning knowledge base~(KB) with an unlabeled corpus. 
It assumes that if an entity pair has a relationship in a KB, all sentences that contain this pair express the corresponding relation.

Nevertheless, DS brings plenty of noisy labels which significantly degrade the performance of the joint extraction models. 
For example, given a sentence ``\emph{Bill Gates lived in Albuquerque}'' and the sentence in Figure~\ref{fig:example}, DS may assign the relation type between ``\emph{Bill Gates}'' and ``\emph{Albuquerque}'' as \emph{Place\_lived} for both sentences. 
The words ``\emph{lived in}'' in the first sentence is the pattern that explains the relation type, thus it is correctly labeled. 
While the second sentence is noisy due to the lack of corresponding relation pattern. 
Moreover, due to the ambiguity and limited coverage over entities in open-domain KBs, DS also generates noisy and incomplete entity labels. 
In some cases, DS may lead to over 30\% noisy instances~\cite{mintz2009distant}, making it impossible to learn useful features.

Previous studies for handling such noisy labels consider either weakly-labeled entities, i.e., distantly-supervised named entity recognition~(NER)~\cite{nadeau2007survey}, or noisy relation labels, i.e., distantly-supervised relation extraction~(RE)~\cite{rink2010utd}, where they focus on designing novel hand-crafted relation features~\cite{yu2019joint}, neural architectures~\cite{chen2020joint}, and tagging scheme~\cite{dai2019joint} to improve relation extraction performance. Additionally, In-Context Learning~(ICL) using external knowledge of Large Language Models~(LLMs)~\cite{pang-etal-2023-guideline} is popular. However, they are resource-demanding, sensitive to prompt design, and may struggle with complex tasks.

To cheaply mitigate both noise sources, we propose \textbf{DENRL}---\textbf{D}istantly-supervised joint \textbf{E}xtraction with \textbf{N}oise-\textbf{R}obust \textbf{L}earning. 
DENRL assumes that 1)~reliable relation labels, whose relation patterns significantly indicate the relationship between entity pairs, should be explained by a model, and 2)~reliable relation labels also implicitly indicate reliable entity tags of the corresponding entity pairs. 
Specifically, DENRL applies \emph{Bag-of-word Regularization}~(BR) to guide a model to attend to significant relation patterns that explain correct relation labels, and \emph{Ontology-based Logic Fusion}~(OLF) that teaches underlying entity-relation dependencies with Probabilistic Soft Logic~(PSL)~\cite{bach2017hinge}. 
These two information sources are integrated to form a noise-robust loss, which regularizes a tagging model to learn from instances with correct entity and relation labels. 
Next, if a learned model clearly locates the relation patterns and understands entity-relation logic of candidate instances, they are selected for subsequent adaptive learning.
We further sample negative instances that contain corresponding head or tail entities of recognized patterns in those candidates to reduce entity noise. 
We iteratively learn an interpretable model and select high-quality instances. 
These two-fold steps are mutually reinforced---a more interpretable model helps select a higher quality subset, and vice versa.

Given the superiority of unified joint extraction methods, we introduce a sequence labeling~\cite{zheng-etal-2017-joint} method to tag entities and their relations simultaneously as token classification. 
We incorporate a BERT~\cite{devlin-etal-2019-bert} backbone that learns rich feature representations into the tagging scheme to benefit the information propagation between relations and entities. 
The transformer attention mechanism builds a direct connection between words and contributes to extracting long-range relations~\cite{li-etal-2022-share,li2023glad}. 
Its multi-head attention weights indicate interactions between each pair of words, which is further leveraged by self-matching to produce position-aware representations. 
These representations are finally used to decode different tagging results and extract all entities together with their relations. 

Our contributions are summarized as follows:
\begin{itemize}[]
    \item Our work introduces a novel framework for distantly-supervised joint extraction. This innovation lies in identifying and addressing multi-source noise arising from both entity and relation annotations, and a unified joint tagging scheme adaptable to various backbones.
    \item Our method DENRL is generalizable and effective and offers a cost-effective alternative to predominant LLMs that use a much larger backbone.
    \item Our comprehensive experiments show that DENRL is interpretible and well-motivated.
\end{itemize}

\section{Joint Extraction Architecture}

\begin{figure}
    \centering
    \includegraphics[width=0.48\textwidth]{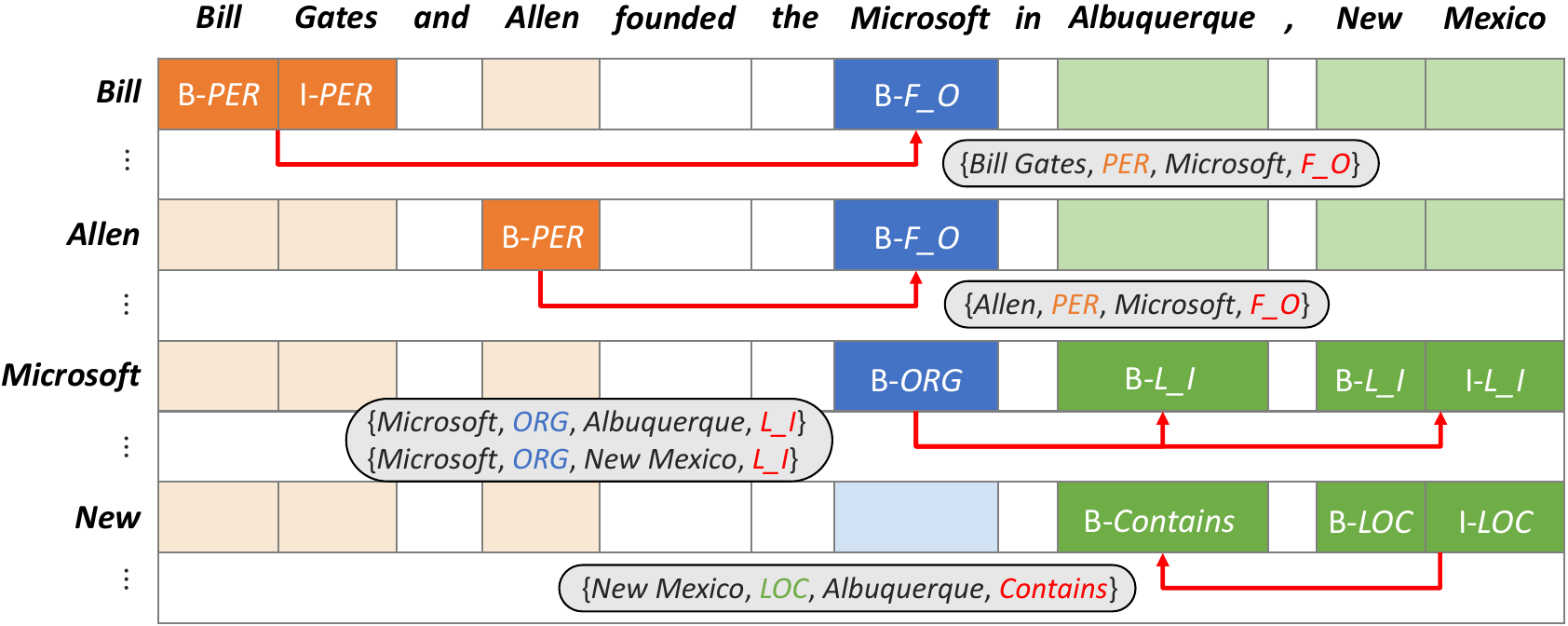}
    \caption{A example of our tagging scheme. For each head entity, we fill a $T$-tag sequence to represent corresponding relations. \emph{PER}, \emph{ORG}, \emph{LOC} are abbreviations for entity \emph{PERSON}, \emph{ORGANIZATION}, \emph{LOCATION}; \emph{F\_O}, \emph{L\_I} for relation \emph{Founder\_of}, \emph{Located\_in}.}
    \label{fig:tag_example}
    \vspace{-0.3cm}
\end{figure}

\begin{figure*}
    \centering
    \includegraphics[width=0.97\textwidth]{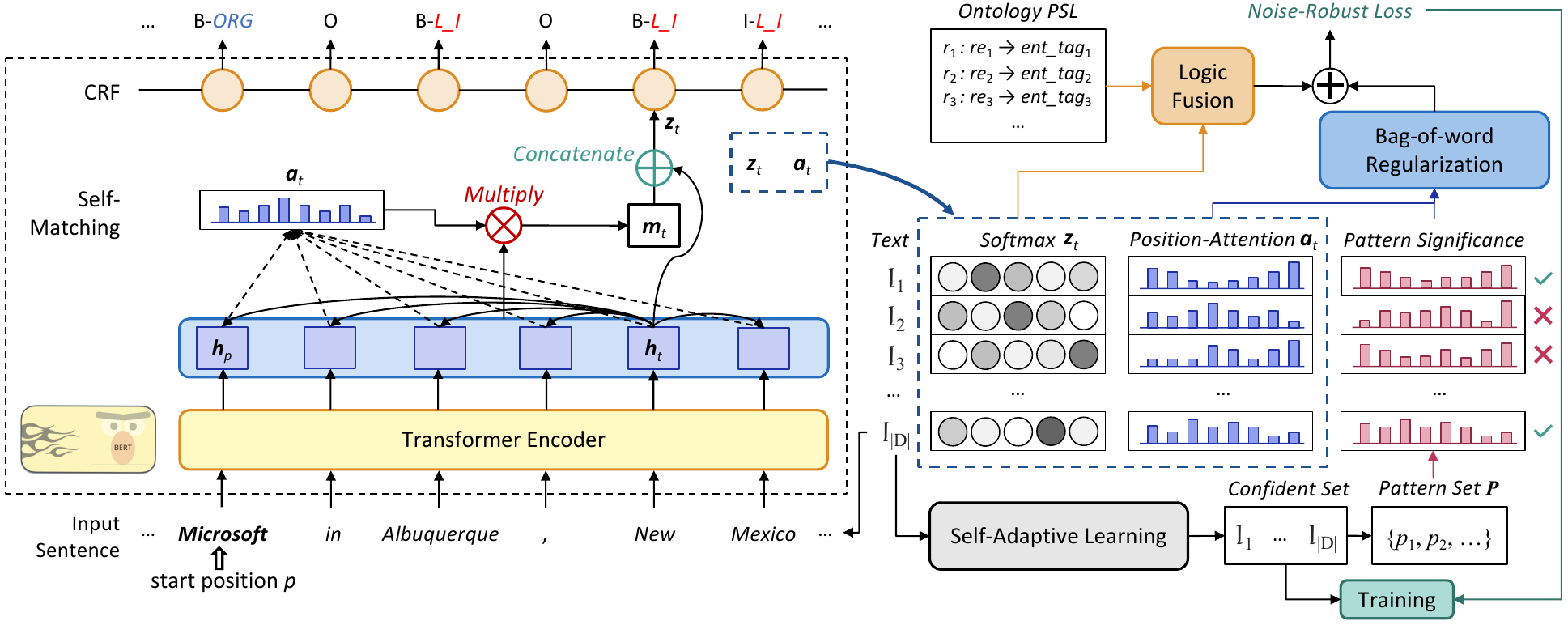}
    \caption{An overview of DENRL framework. 
    The left part is our position-attentive joint tagging model, which receives a sentence input and different start position $p$ to extract all entities and relations. 
    $\bm{a}_t$ are position-attention weights and $\bm{z}_t$ are sequence scores. 
    The right part is our noise-robust learning mechanism, which employs BR (on $\bm{a}_t$) and OLF (on $\bm{z}_t$) to guide the model to attend to significant patterns and entity-relation dependencies. 
    Then, a fitness score $u$ for each training instance is calculated to select and build new distributed training sets as well as confident pattern sets. 
    These two steps are run iteratively as self-adaptive learning.}
    \label{fig:architecture}
    \vspace{-0.3cm}
\end{figure*}

We incorporate a pre-trained BERT backbone into our sequence tagging scheme to jointly extract entities and their relations (see Figure~\ref{fig:architecture}).

\subsection{Tagging Scheme}

To extract both entities (mention and type) and relations, we tag quadruplets $\left \{ e_1, tag_1, e_2, re \right \}$ for each start position $p$ and define ``BIO'' signs to encode positions (see Figure~\ref{fig:tag_example}). 
Here, $e_1$ is the detected entity at $p$~(head entity), $tag_1$ is the entity type of $e_1$, $e_2$ is another detected entity that has a relationship with $e_1$~(tail entity), and $re$ is the predicted relation type between $e_1$ and $e_2$.
For a $T$-token sentence, we annotate $T$ different tag sequences according to different start positions. 

For each tag sequence, if $p$ is the start of an entity (this sequence is an instance), the entity type is labeled at $p$, and other entities that have relationship to the entity at $p$ are labeled with relation types. 
The rest of tokens are labeled ``O''~(Outside), meaning they do not correspond to the head entity.
In this way, each tag sequence will produce a relation quadruplet. 
For example, if $p$ is 7, the head entity is ``\emph{Microsoft}'' and its tag is \emph{ORG}.
Other entities, such as ``\emph{Albuquerque}'' and ``\emph{New Mexico}'', are labeled as \emph{L\_I} and \emph{L\_I} indicating their (unidirectional) relations with ``\emph{Microsoft}''. 
If $p$ is 9, the head entity ``\emph{Albuquerque}'' has no relationship with other entities, thus only the entity type \emph{LOC} is labeled. 
If $p$ is 13, all tokens are labeled as ``O'' because there is no entity at the head position to attend to.

We define instances that contain at least one relation as positive instances (e.g., $p$ is 7), and those without relations as negative instances (e.g., $p$ is 9). ``BIO''~(Begin, Inside, Outside) signs are used to indicate the position information of tokens in each entity for both entity and relation type annotation to extract multi-word entities. 
Note that we do not need the tail entity type, because every entity will be queried and we can obtain all entity types as well as their relations from the $T$ tag sequences.

\subsection{Tagging Model}

\textbf{Encoder with Self-Matching}
We follow BERT~\cite{devlin-etal-2019-bert} to use a multilayer transformer~\cite{vaswani2017attention} that takes an input sequence $\mathcal{S} = \left \{ w_1,...w_T \right \}$ and converts it into token-level representations $\bm{h}^{0} = \left \{ \bm{h}_{t} \right \}_{t=1}^{T}$, where $\bm{h}_{t} \in \mathbb{R}^{d}$ is a $d$-dimensional vector corresponding to the $t$-th token in $\mathcal{S}$. 
The model applies $L$ transformer layers over the hidden vectors to produce contextual representations:
$\bm{h}^{l} = \textsc{Transformer}^{(l)}(\bm{h}^{l-1}),~~l\in [1,L]$.
Each layer contains a Multi-Head Self-Attention~(MHSA) layer followed by a Feed-Forward Network~(FFN) over the previous hidden state $\bm{h}^{l-1}$.
The final representations $\bm{h}^{L}\in \mathbb{R}^{T\times d}$ integrate the contextual information of all previous tokens but are inadequate for decoding a $T$-tag sequence, since for each position $p$ we still need to encode $e_1$ and its overlapping relations $re$ with other entities $e_2$.

We define \emph{self-matching}~\cite{tan2018deep} that calculates position-attention $\bm{a}_{t}$ between tokens at start position $p$ as well as each target position $t$:
\begin{equation}
\label{eq: position_attention}
\begin{gathered}
    \bm{a}_{t} = \text{softmax} (\left \{ a^t_j \right \}_{j=1}^{T}) \\
    s.t. \;\; a^t_j = \bm{w}^{\top}(\bm{h}_{p}^{L}+\bm{h}_{t}^{L}+\bm{h}_{j}^{L}),
\end{gathered}
\end{equation}
where $\bm{w}\in \mathbb{R}^{d}$ is a parameter to be learned, $\bm{h}_{p}$, $\bm{h}_{t}$, $\bm{h}_{j} \in \mathbb{R}^{d}$ are hidden states at position $p$, $t$, $j$, respectively. 
$a^t_j$ is the score computed by comparing $\bm{h}_{p}$ and $\bm{h}_{t}$ with each hidden state $\bm{h}_{j}$. 
$\bm{a}_{t} \in \mathbb{R}^{T}$ is the softmax attention produced by normalizing $a^t_j$.
The start hidden state $\bm{h}_{p}$ serves as comparing with the sentence representations to encode position information, and $\bm{h}_{t}$ matches the sentence representations against itself to collect context information. 
The position-aware representation $\bm{m}_{t}\in \mathbb{R}^{T\times d}$ is an attention-weighted sentence vector: 
\begin{equation}
  \bm{m}_{t} = \bm{a}_{t}^{\top} \bm{h}^{L}.
\end{equation}
We concatenate $\bm{h}_t$ and $\bm{m}_{t}$ to generate position-aware and context-aware representations $\left \{ \bm{x}_{t} \right \}_{t=1}^{T}$: 
\begin{equation}
    \bm{x}_{t} = [\bm{h}_t; \bm{m}_{t}].
\end{equation}
For each start position, self-matching produces different sentence representations and thus can model different tag sequences of a sentence.

\textbf{CRF Decoder} CRF~\cite{lafferty2001conditional} considers the correlations between labels in neighborhoods and jointly decodes the best chain of labels, which benefits sequence labeling models. 
For each position-aware representation $\bm{x}_t$,  the input sequence scores $\bm{Z} = \left \{\bm{z}_{t} \right \}_{t=1}^{T}$ is generated by:
\begin{equation}
    \label{eq:z_t}
    \bm{z}_{t} = \bm{W}^{x}\bm{x}_{t},
\end{equation}
where $\bm{z}_{t} \in \mathbb{R}^{V}$ is tag score of the $t$-th token, $V$ is the number of distinct tags, and $z_{t}^j$ is the score of the $j$-th tag at position $t$. 

For a sequence of labels $\bm{y} = \left \{ y_1,...,y_T \right \}$, the decoding score $score(\bm{Z}, \bm{y})$ is the sum of transition score from tag $y_t$ to tag $y_{t+1}$, plus the input score $z_t^{y_t}$ for each token position $t$. 
The conditional probability $p(\bm{y}|\bm{Z})$ is the softmax of $score(\bm{Z}, \bm{y})$ over all possible label sequences $\bm{y}'$ for $\bm{Z}$. We maximize the log-likelihood of correct tag sequences during training: 

\begin{equation}
    \label{eq:loss_c}
    \mathcal{L}_c = \sum\nolimits_{i}^{}{\log p(\bm{y}|\bm{Z})}.
\end{equation}

Decoding searches for the tag sequence $\bm{y}^{*}$ that maximizes the decoding score.
The best tag sequence $\bm{y}^{*}$ is computed using the Viterbi algorithm.

\section{Noise-Robust Learning}
\label{sec:denoise}

To reduce the impact of noisy labels on tagging performance, we introduce \emph{Bow Regularization}~(BR) to attend to confident relation patterns for reducing relation noise and \emph{Ontology-based Logic Fusion}~(OLF) to increase entity-relation coherence for reducing entity noise. 
Finally, we employ \emph{Self-Adaptive Learning}~(SAL) to iteratively train on instances that can be explained by the model.   

\subsection{Bag-of-word Regularization~(BR)} 
Originally proposed as a pattern-attentive loss, attention regularization~\cite{jia-etal-2019-arnor} has been shown effective for reducing relation noise, yet it only considers attention over a single relation pattern and neglects models' position-awareness, thus cannot identify overlapping relations.
We formulate target attention distribution by introducing BoW frequency as an oracle to learn informative relations.
For an input sentence $\mathcal{S}$, an entity pair $(e_1, e_2)$ in $\mathcal{S}$, a relation label $re$, and a relation pattern $p$ that explains the relation $re$ of $e_{1}$ and $e_{2}$, we define BoW frequency (i.e., pattern significance) as the corresponding guidance score $\bm{a}^p$ conditional on pattern $p$. 
Take the relation \emph{Contains} as an example, its BoW is a set of tokens that appear in a corresponding pattern set $\left \{\emph{``capital of'', ``section in'', ``areas of'', ...}\right \}$.
The motivation is to guide the model to explore new high-quality patterns $p$, e.g., ``\emph{section of}'', ``\emph{areas in}'', etc.
The guidance $\bm{a}^{\mathcal{I}}$ for an instance $\mathcal{I}$ is the average of $\bm{a}^p$ regarding all patterns $m$ corresponding to each relation type $re$:
\begin{equation}
\label{eq: target_attention}
  \begin{gathered}
  \bm{a}^p =\text{softmax}(\left \{ \text{BoW}_t \right \}_{t=1}^{T}), \\
  \bm{a}^{\mathcal{I}} = \text{AvgPooling}\left ( \bm{a}^{p_1},\cdots, \bm{a}^{p_{|R_{\mathcal{I}}|}} \right ),
  \end{gathered}
\end{equation}
where $BoW_t$ represents the BoW frequency of $w_t$ under relation $re$ if $w_t$ belongs to corresponding relation pattern words or 1 if it belongs to entity words, e.g., $f(\emph{``of''}|\emph{Contains})=2$. $|R_{\mathcal{I}}|$ is the number of distinct relation types in instance $\mathcal{I}$.

We expect a joint tagger to approximate its position-attention $\bm{a}^{\mathcal{S}}$ to $\bm{a}^{\mathcal{I}}$, where $\bm{a}^{\mathcal{S}} = \text{AvgPooling}\left (\bm{a}_{1}, \dots, \bm{a}_{T}\right )$ is the average pooling of model's position-attention $\bm{a}_t$ defined in Equation~(\ref{eq: position_attention}) for each position $j$ in ${\mathcal{S}}$. 
We compute the mean squared error (MSE) as the objective:
\begin{equation}
\label{eq:loss_r}
    \mathcal{L}_{BR} = \text{MSE}(\bm{a}^{\mathcal{I}}, \bm{a}^{\mathcal{S}}) = \sum{(\bm{a}^{\mathcal{I}} - \bm{a}^{\mathcal{S}})^2}.
\end{equation}

\subsection{Ontology-Based Logic Fusion~(OLF)} Probabilistic Soft Logic~(PSL)~\cite{bach2017hinge} uses soft truth values for predicates in an interval between $[0,1]$, which represents our token classification probability $p(y_t|w_t)$ as a convex optimization problem.
Inspired by \citet{wang2020integrating,kirsch2020noise} that considers relation logic as rules for inference, we adapt PSL to entity-relation dependency rules according to data ontology using human annotation (see details in Table~\ref{tab:logic_NYT} and Table~\ref{tab:logic_KBP}), e.g., relation type \emph{Founder\_of} should entail (head) entity type \emph{PERSON}. Training instances that violate any of these rules are penalized to enhance comprehension of entity-relation coherence. Suppose BR guides a model to recognize confident relations, OLF further helps explore instances with reliable entity labels, especially when no relations exist in them. 

Particularly, we define \emph{Logic Distance} based on a model's softmax scores over the head entity given its predicted relation type to measure how severely it violates logic rules.
For a training instance, we define an \emph{atom} $l$ as each tag and the \emph{interpretation} $I(l)$ as the soft truth value for the atom. 
For each rule $r: \textsc{relation} \rightarrow \textsc{entity}$, the distance to satisfaction $d_r(I)$ under the interpretation $I$ is:
\begin{equation}
    \label{eq:distance}
    d_r(I) = \max{\left \{0, I(l_{re} )-I(l_{ent}) \right \}}.
\end{equation}
PSL determines a rule $r$ as satisfied when the truth value of $I(l_{re})-I(l_{ent}) \geq 0$. For each instance $\mathcal{I}$, we set $l_{ent}$ as (head) entity type and $l_{re}$ as relation type.
This equation indicates that the smaller $I(l_{ent})$ is, the larger the penalty it has. We compute the distance to satisfaction for each rule $r$ and use the smallest one as a penalty because at least one rule needs to be satisfied.

We learn a distance function $\mathcal{D}(\cdot,\cdot)$ that minimizes all possible PSL rule grounding results, as described in Algorithm~\ref{alg:logic}.
$\mathcal{D}(\cdot,\cdot)$ should return 0 if at least one PSL rule is satisfied. 
The prediction probability $p(y|e_1)$ over head entity $e_1$ is regarded as the interpretation $I(l_{ent})$ of ground atom $l_{ent}$, so as $p(y|e_2)$ over tail entity $e_2$ for $I(l_{re})$ of $l_{re}$. 
If no rules are satisfied, the distance is set as 0. We formulate the distance to satisfaction as a regularization term to penalize inconsistent predictions:
\vspace{-0.3cm}
\begin{equation}
    \label{eq:loss_p}
    \mathcal{L}_{OLF} = \sum{\mathcal{D}(\mathcal{R};\left \{ (p(y|e_i),\hat{y}_i) \right \})},
\end{equation}
where $p(y|e_i)$ is the softmax probability of $\bm{z}_{t_i}$ in Equation~(\ref{eq:z_t}) for position $t_i$ of $e_i$ in $\mathcal{S}$, and $\mathcal{L}_{OLF}$ is the sum of $\mathcal{D}(\cdot,\cdot)$ over all entity-relation pairs $(e_1, e_2)$ in instance $\mathcal{I}$.
We finalize a noise-robust loss function by summing up (\ref{eq:loss_c}), (\ref{eq:loss_r}) and (\ref{eq:loss_p}):
\begin{equation}
    \mathcal{L} = \mathcal{L}_c + \alpha \mathcal{L}_{BR} + \beta \mathcal{L}_{OLF},
\end{equation}
where $\alpha$, $\beta$ are two balancing hyper-parameters.

\begin{algorithm}[t]
\caption{Logic Distance Calculation $\mathcal{D}$}\label{alg:logic}
\begin{algorithmic}[1]
\Require Softmax $p(y|e_i)$, Prediction $\hat{y}_i$, $i\in \left \{1,2 \right \}$, PSL rules $\mathcal{R}$ w.r.t. ontology;
\Ensure Distance $d$;
\State Initialize $d \leftarrow 1$; Satisfied $\leftarrow$ False;
\For {each $r: l_{re} \rightarrow l_{ent} \in \mathcal{R} \wedge \hat{y}_{2} == l_{re} $}
    \State $\overline{y}_{1} \leftarrow l_{ent}$;
    \State $d' \leftarrow \max{\left \{ p(\hat{y}_{2}|e_{2}) - p(\overline{y}_{1}|e_{1}), 0 \right \}}$;
    \State $d \leftarrow \min{\left \{ d', d \right \}}$;
    \State Satisfied $\leftarrow$ True;
\EndFor
\If {Satisfied $==$ False} \State $d \leftarrow 0$. 
\EndIf
\end{algorithmic}
\end{algorithm}

\subsection{Self-Adaptive Learning (SAL)} 
Self-adaptive learning~\cite{jia-etal-2019-arnor} aims to iteratively select high-quality instances with informative relation patterns $p$ and entity tags. 
In each training epoch, more precisely labeled instances are chosen to guide a model to attend to informative evidence for joint extraction. 
For instance selection, more versatile patterns are required to select clean labels and to discover more confident relation patterns. 
According to the attention mechanism and entity-relation logic, a trained tagger can tell the importance of each word for identifying the entity pair along with their relationship, and predict reasonable entity-relation label pairs. 
For instance $\mathcal{I}$, if 1) the model's attention weights do not match the target attention that explains the relation types in $\mathcal{I}$, or 2) its confidence distribution over entity and relation tags violates the logic dependencies, this instance is likely a false alarm. 
We add up both BR and OLF loss for an instance $\mathcal{I}$ to measure its \emph{fitness} $u(\mathcal{I})$, i.e., how likely it is correctly labeled: 
\begin{equation}
    u = \sigma[\text{MSE}(\bm{a}^{\mathcal{I}}, \bm{a}^{\mathcal{S}})+\mathcal{D}(\mathcal{R};\mathcal{I})],
\end{equation}
where $\sigma$ is the sigmoid function that bounds $u$ in the range $[0, 1]$.
The lower $u$ is, the more confident an instance $\mathcal{I}$ is. 
We compute fitness scores for all training instances and select those whose score is smaller than a predefined threshold $\tau$. 

Because trustable relation labels also indicate trustable entity tags, we further consider \emph{Entity Selection}~(ES), i.e., selecting negative instances containing either the head or tail entity corresponding to each relation pattern in the selected positive candidates.
Specifically, we consider relation pattern $p$ as the text between two entities in an instance. 
We build an initial trustable pattern set $\mathcal{P}$ by counting all patterns up and selecting the top 10\% frequent patterns for each relation type. 
Next, we redistribute the training dataset $\mathbf{D}$ based on $\mathcal{P}$, where all positive instances that match patterns in $\mathcal{P}$ as well as negative instances that contain the head entity or tail entity of these patterns are retained to train the model for a few epochs. 
Finally, we select more reliable instances according to fitness scores over $\mathbf{D}$, from which we extract new trustable patterns to enrich $\mathcal{P}$. 
These new confident instances are learned in the subsequent iteration. 
We repeat the above procedure until the validation F1 converges. 



\section{Experiments}

\begin{table*}[t]
    \centering
    \resizebox{\textwidth}{!}{
    \begin{tabular}{l|ccc|ccc}
        \hline
        \multirow{2}{*}{\textbf{Method}} & \multicolumn{3}{c|}{\textbf{NYT}} & \multicolumn{3}{c}{\textbf{Wiki-KBP}}\\
        \cline{2-7}
        
        &\textbf{Prec.} & \textbf{Rec.} & \textbf{F1}&\textbf{Prec.} & \textbf{Rec.} & \textbf{F1}  \\
        \hline
        LSTM-CRF~\cite{zheng-etal-2017-joint} & 66.73 & 35.02 & 45.93 & 40.14 & 35.27 & 37.55 \\
        PA-LSTM-CRF~\cite{dai2019joint} & 37.90 & \textbf{76.25} & 50.63 & 35.82 & 45.06 & 39.91 \\
        OneIE~\cite{lin-etal-2020-joint} & 52.33 & 64.40 & 57.74 & 36.25 & 46.51 & 40.74 \\
        PURE~\cite{zhong-chen-2021-frustratingly} & 53.11 & 65.84 & 58.79 & 38.20 & 44.89 & 41.28 \\
        \cline{1-7}
        CoType~\cite{ren2017cotype} & 51.17 & 55.92 & 53.44 & 35.68 & 46.39 & 40.34 \\
        CNN+RL~\cite{feng2018reinforcement} & 40.72 & 58.39 & 47.98 & 36.20 & 44.57 & 39.95 \\
        PCNN+RL~\cite{qin-etal-2018-robust} & 46.84	& 53.15	& 49.80 & 37.75 & 42.36 & 39.92 \\
        ARNOR~\cite{jia2019arnor} & 59.64 & 60.78 & 60.20 & 39.37 & 47.13 & 42.90 \\
        FAN~\cite{hao-etal-2021-knowing} & 58.22 & 64.16 & 61.05 & 38.81 & 47.14 & 42.57 \\
        SENT~\cite{ma-etal-2021-sent} & 63.88 & 62.12 & 62.99 & 41.37 & 46.72 & 43.88 \\
        \hline
        Llama-ICL~\cite{pang-etal-2023-guideline} & 61.81 & 58.79 & 60.26 & 40.52 & 45.60 & 42.91 \\
        GPT-4-ICL~\cite{pang-etal-2023-guideline} & 63.04 & 57.69 & 60.25 & 44.14 & 41.92 & 43.00 \\
        \hline
        \textbf{DENRL}~(triplet) & \textbf{69.37}$_{\pm0.68}$ &	67.01$_{\pm0.70}$ & \textbf{68.17}$_{\pm0.69}$ &	\textbf{42.49}$_{\pm0.31}$ &	\textbf{50.78}$_{\pm0.25}$ & \textbf{46.27}$_{\pm0.28}$ \\
        \textbf{DENRL} & 69.24$_{\pm0.61}$ &	66.23$_{\pm0.44}$ &	67.70$_{\pm0.52}$ &	41.96$_{\pm0.34}$ &	50.21$_{\pm0.26}$ &	45.72$_{\pm0.30}$ \\
        \hline
        
    \end{tabular}}
    \caption{Evaluation results on NYT and Wiki-KBP datasets.
    Baselines include normal RE methods (the 1st part), DS RE methods (the 2nd part), and ICL method (the 3rd part).
    We ran the model 5 times to get the average results.
    }
    \label{tab:main_results}
    \vspace{-0.3cm}
\end{table*}

\subsection{Datasets and Evaluation}
We evaluate the performance of DENRL on two public datasets: (1)~\textbf{NYT}~\cite{riedel2010modeling}. We use the human-annotated test dataset~\cite{jia2019arnor} including 1,024 sentences with 3,280 instances and 3,880 quadruplets. 
The training data is automatically generated by DS (aligning entity pairs from Freebase with handcrafted rules), including 235$k$ sentences with 692$k$ instances and 353$k$ quadruplets. 
(2)~\textbf{Wiki-KBP}~\cite{ling2012fine}. Its test set is manually annotated in 2013 KBP slot filling assessment results~\cite{KBP_slot} containing 289 sentences with 919 instances and 1092 quadruplets. The training data is generated by DS~\cite{liu-etal-2017-heterogeneous} including 75k sentences with 145k instances and 115k quadruplets.

We evaluate the extracted quadruplets for each sentence in terms of Precision (Prec.), Recall (Rec.), and F1. 
A quadruplet $\left \{ e_1, tag_1, e_2, re \right \}$ is marked correct if the relation type $re$, two entities $e_1$, $e_2$, and head entity type $tag_1$ are all matched. 
Note that negative quadruplets with ``None'' relation are also considered for evaluating prediction accuracy. 
We build a validation set by randomly sampling 10\% sentences from the test set. 

 

\subsection{Baselines}
We compare DENRL with the following baselines:
    
\noindent\textbf{LSTM-CRF}~\cite{zheng-etal-2017-joint} that converts joint extraction to a sequence labeling problem based on a novel tagging scheme.

\noindent\textbf{PA-LSTM-CRF}~\cite{dai2019joint}, which uses sequence tagging to jointly extract entities and overlapping relations.

\noindent\textbf{OneIE}~\cite{lin-etal-2020-joint}, a table-filling approach that uses an RNN table encoder to learn sequence features for NER and a pre-trained BERT sequence encoder to learn table features for RE. 

\noindent\textbf{PURE}~\cite{zhong-chen-2021-frustratingly}, a pipeline approach that uses a pre-trained BERT entity model to first recognize entities and then employs a relation model to detect underlying relations.

\noindent\textbf{CoType}~\cite{ren2017cotype}, a feature-based method that handles noisy labels based on multi-instance learning, assuming at least one mention is correct.

\noindent\textbf{CNN+RL}~\cite{feng2018reinforcement} that trains an instance selector and a CNN classifier using reinforcement learning.

\noindent\textbf{PCNN+RL}~\cite{qin-etal-2018-robust}, a baseline whose RL method used to detect and remove noise instances is independent of the training of RE systems.

\noindent\textbf{ARNOR}~\cite{jia2019arnor} which uses attention regularization and bootstrap learning to reduce noise for DS RE.

\noindent\textbf{FAN}~\cite{hao-etal-2021-knowing}, an adversarial method including a BERT encoder to reduce noise for DS RE.

\noindent\textbf{SENT}~\cite{ma-etal-2021-sent}, a negative training method that selects complementary labels and re-labels noisy instances with BERT for DS RE.

\noindent\textbf{Llama-ICL}~\cite{pang-etal-2023-guideline}, we follow the basic prompt with two demonstration examples using Llama, each as a pair of input text and extracted triplets. 

\noindent\textbf{GPT-4-ICL}~\cite{pang-etal-2023-guideline}, the same setting as Llama-ICL but with GPT-4 as the backbone.

\subsection{Implementation Details}

For DENRL and baselines using pre-trained BERT, we use the pre-trained \emph{bert-large-cased} from Hugging Face.
For baselines using LSTM, we apply a single layer with a hidden size of 256.
We also extend DENRL to two other backbones: T5 and GPT-2, and demonstrate the generalizability of DENRL in Appendix~\ref{sec:backbone_generalizability}.
For Llama-ICL, we use Llama-2-7B~\cite{touvron2023llama}.
The prompt configuration and training setup for both Llama-2 and GPT-4 are detailed in Appendix~\ref{sec:implementation}.

\subsection{Overall Results}
As shown in Table~\ref{tab:main_results}, DENRL~(triplet) denotes ignoring head entity type $tag_1$ when computing correctness, because all baselines only extract triplets $\left \{ e_1, e_2, re \right \}$. 
The results of triplet and quadruplet have little difference, indicating that DENRL predicts precise entity types. 
DENRL significantly outperforms all baselines in precision and F1 metrics. Specifically, it achieves roughly 5$\sim$20\% F1 improvement on NYT (3$\sim$6\% on Wiki-KBP) over the other denoising methods---CoType, CNN+RL, ARNOR, FAN, SENT. 
Compared to LSTM-CRF which also trains on selected subsets, DENRL achieves 31\% recall improvements on NYT (15\% on Wiki-KBP) with still better precision, suggesting that we explore more diverse entity and relation patterns. 
Compared to the sequence tagging approach PA-LSTM-CRF, DENRL achieves improvements of 32\% in precision and over 18\% F1 improvement. 
DENRL also outperforms baselines using pre-trained transformers (OneIE, PURE, FAN, SENT) or LLMs (Llama-2-ICL, GPT-4-ICL), showing our noise-robust learning effectively reduces the impact of mislabeled instances on joint extraction performance.

\subsection{Ablation Study}
\begin{table}[t]
    \centering
    \resizebox{0.47\textwidth}{!}{
    \begin{tabular}{lccc}
        \hline
        \textbf{Component} & \textbf{Prec.} & \textbf{Rec.} & \textbf{F1} \\
        \hline
        BERT+FC & 44.17 & 72.76 & 54.97 \\ 
         BERT+CRF & 44.98 & \textbf{74.79} & 56.18 \\ 
         \hline
         +IDR & \textbf{73.18} & 48.60 & 58.41 \\ 
         +BR & 69.33 & 54.02 & 60.72 \\
         +OLF & 70.89 & 52.14 & 60.09 \\
         +BR+OLF & 71.35 &  56.29 & 62.93 \\
         +BR+SAL & 70.62 &  62.84 &	66.50 \\
         +OLF+SAL & 70.81 & 59.76 & 64.82 \\
         +BR+OLF+SAL (DENRL) & 69.37 & 67.01 & \textbf{68.17} \\
         \hline
         
    \end{tabular}}
    \caption{Ablation study of components in DENRL. 
    BERT+FC and BERT+CRF are two backbone models. 
    IDR denotes initial data redistributing using the initial pattern set. 
    BR and OLF are only for the first loop.}
    \label{tab:ablation_study}
    \vspace{-0.3cm}
\end{table}

We investigate the effectiveness of several components of DENRL on NYT dataset, as shown in Table~\ref{tab:ablation_study}. 
Before noise reduction, we first evaluate the impact of CRF layer by substituting it with an FC layer. 
We found it improves the final performance by over 1\% F1.
We then build an initial redistributed dataset~(via IDR), which helps the joint model earn over 2\% improvement in F1 and a sharp 28\% precision increase compared to BERT+CRF. 
This suggests the original DS dataset contains plenty of noise, thus a simple filtering method would effectively improve the performance. 

However, this initial data induces poor recall performance, which means a large proportion of true positives with long-tail patterns are mistakenly regarded as false negatives. 
Assuming that some relation patterns in the training data are too rare to guide the model to learn to attend them, we employ BR to training and achieve 5\% recall increases with a slight decline in precision, inducing another 2\% F1 improvement. 
This shows the effect of guiding the model to understand important feature words for identifying relations. 

After we introduce OLF to training, both precision and recall improve by about 2\%, leading to another 2\% F1 improvement, proving that logic rules guide a model to learn the entity-relation dependencies and further reduce entity labeling noise. 

After we obtain an initial model trained by BR and OLF, we continue SAL where DENRL collects more confident long-tail patterns to mitigate false negatives and finally achieves 5\% F1 improvement. 
We observe that BR better facilitates SAL than OLF regarding the recall increase, as BoW helps explore more high-quality relation instances and reduces false negatives. Additionally, BR+OLF outperforms both BR and OLF. This proves our assumption that significant relation patterns and entity-relation dependencies can reinforce each other during training, i.e., understanding significant relation patterns facilitates the predictions of correct entity-relation dependencies, and vice versa.


\begin{table}[t]
    \centering
    \resizebox{0.32\textwidth}{!}{
    \begin{tabular}{lccc}
    \hline
    \textbf{Method} & \textbf{Prec.} & \textbf{Rec.} & \textbf{F1} \\
    \hline
    w/o ES & 67.39 & \textbf{67.04} & 67.21 \\
    DENRL  & \textbf{69.37}  & 67.01 & \textbf{68.17} \\
    \hline
    \end{tabular}}
    \caption{Comparison of Precision, Recall, and F1 after using Entity Selection~(ES) during SAL.}
    \label{tab:entity_selection}
    \vspace{-0.3cm}
\end{table}

\subsection{Interpretability Study}
\label{sec:case study}

\begin{figure*}[hpt]
     \centering
     \begin{subfigure}[h]{0.52\textwidth}
     \centering
     \includegraphics[ height=4.3cm]{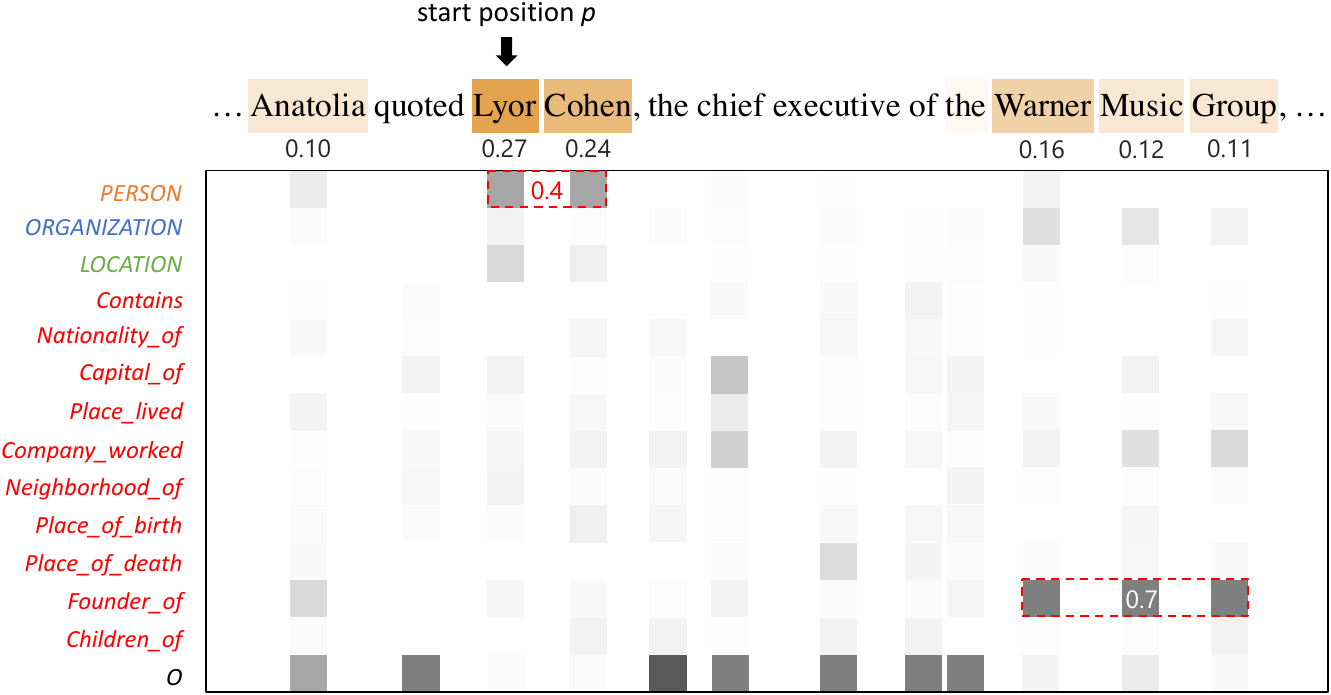}
     \caption{BERT+CRF}
     \label{fig:BERT-PA-CRF attention}
     \end{subfigure}
     \hfill
     \begin{subfigure}[h]{0.47\textwidth}
     \centering
     \includegraphics[height=4.3cm]{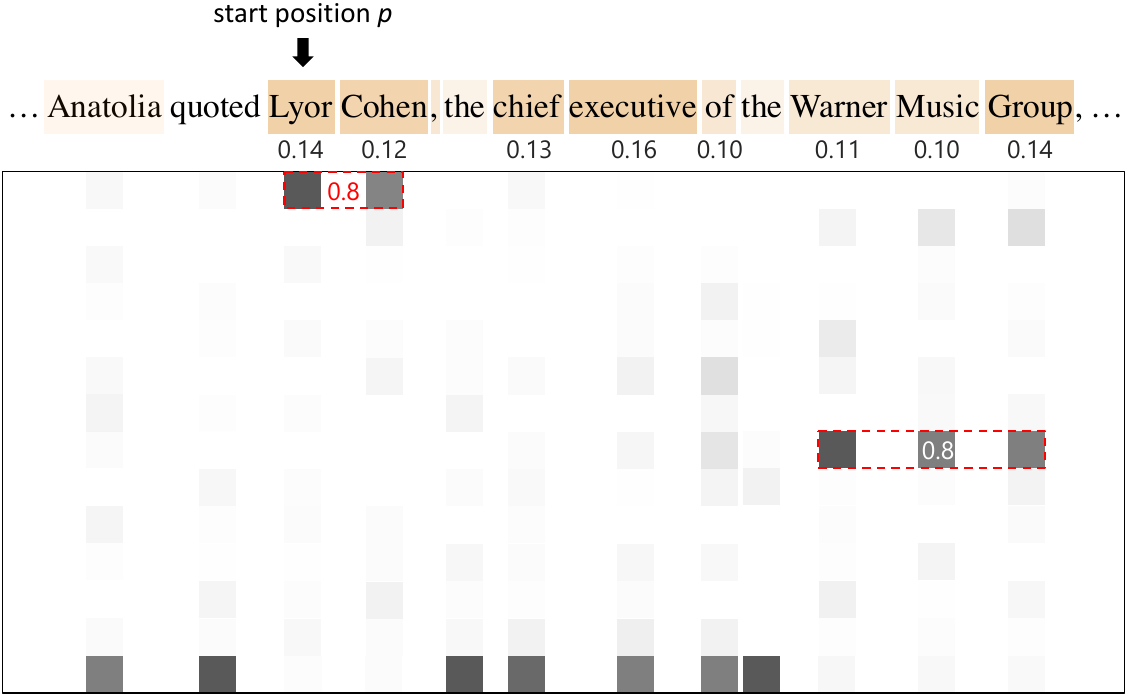}
     \caption{DENRL}
     \label{fig:DENRL attention}
     \end{subfigure}
     
    \caption{Attention heat maps (top) and softmax probability heat maps (bottom). In this case, $e_1$: Lyor Cohen, $e_2$: Warner Music Group, and $re$: \emph{Company\_worked}. BERT+CRF misclassifies the relation as \emph{Founder\_of}, because it only attends to entities. DENRL can locate relation indicators and make correct predictions.}
    \label{fig:heat_map}
    \vspace{-0.3cm}
\end{figure*}

To understand the effect of attention and logic guidance, we select some instances from the test set and visualize their attention weights, as well as the model's softmax probability distribution over all labels. As shown in Figure~\ref{fig:heat_map}, BERT+CRF, which is trained on original noisy data without BR or OLF, only focuses on entity pairs and makes wrong predictions. Its logic distance for $r: \emph{Founder\_of} \rightarrow \emph{PERSON}$ is $d_r(I) = \max{\left \{0, 0.7-0.4 \right \}}=0.3$. While DENRL precisely captures important words and correctly predicts the relation. The logic distance for $r: \emph{Company\_worked} \rightarrow \emph{PERSON}$ is $d_r(I) = \max{\left \{0, 0.8-0.8 \right \}}=0 < 0.3$, suggesting the effect of OLF.

Suppose the start position is at ``Lyor'', for a target position at ``Warner'', the model should pay more attention to (1) the pattern words ``chief executive of'' to determine their correct relation type \emph{Company\_worked}, and (2) the entity words ``Lyor Cohen'' to determine the correct head entity type. This is exactly how attention guidance is built in BR. During BR, for the relation \emph{Company\_worked}, its pattern set contains corresponding patterns, e.g., $\left \{ \emph{`CEO at’, `manager in’, `chief executive of’, ...}\right \}$ and the BoW will count tokens and their frequencies, e.g., $\left \{ \emph{`chief’: 8, `manager’: 14, `at’: 7, `of’: 16, }\right \}$. Consequently, the guidance scores for the words ``chief”, ``executive”, and ``of” become large (after softmax computation) as they appear in the BoW with high frequencies. Also, the head entity tokens ``Lyor'', and ``Cohen'' will be given a score of 1. In this way, both ``Lyor Cohen'' and ``chief executive of'' will be assigned a high guidance score, and we want the model’s attention to approximate such guidance distribution (by minimizing their discrepancy) so that it can better identify correct relations.


We further check the performance of DENRL on negative test cases that do not contain relations from NYT dataset. 
After selecting confident candidates in each epoch, we further choose additional trustable negative instances that contain either the head or tail entity corresponding to each relation pattern in the selected positive candidates during bootstrap. 
We compare the results between methods with and without entity selection, as shown in Table~\ref{tab:entity_selection}. 
The improved performance with ES demonstrates that a trustable relation pattern also indicates reliable entity labels, and partially explains the overall superiority of DENRL.

\subsection{Efficiency Analysis}
\label{sec:efficiency}

\begin{table}[]
    \centering
    \resizebox{0.35\textwidth}{!}{
    \begin{tabular}{c|cc}
    \hline
        \textbf{Method} & \textbf{NYT} & \textbf{Wiki-KBP}  \\
    \hline
        BERT+CRF & 0.78 & 0.70 \\
        T5+CRF & 0.89 & 0.82 \\
        GPT-2+CRF & 0.94 & 0.88 \\
    \hline
        LSTM-CRF & \textbf{0.27} & \textbf{0.21} \\
        PA-LSTM-CRF & 0.35 & 0.33 \\
        OneIE & 0.32 & 0.28 \\
        PURE & 0.85 & 0.79 \\
    \hline
        ARNOR & 0.43 & 0.39 \\
        FAN & \underline{1.62} & \underline{1.59} \\
        SENT & 1.43 & 1.36 \\
    \hline
        DENRL & 1.39 & 1.07 \\
    \hline
    \end{tabular}}
    \caption{Comparison of training efficiency (GPU hours) between baselines and SAL training of DENRL with different backbones. \textbf{Bold} and \underline{underline} denote most efficient and time-consuming methods.}
    \label{tab:efficiency}
    \vspace{-0.3cm}
\end{table}

While DENRL considers the position-attentive loss calculated through traversing transformer logits on different start positions, it does not significantly inflate the training overhead. 
For a sentence of $n$ tokens, the time-intensive self-attention operations (squared complexity) are executed just once per sentence. 
The resultant hidden outputs are used to perform self-matching and CRF decoding regarding each start token, which also has an $O(n^2)$ complexity but with few extra trainable parameters introduced. 
This layered approach ensures a manageable overall computational overhead.
Table~\ref{tab:efficiency} shows the average GPU hours per training epoch for each method. We observe that DS methods consume more time compared to their normal counterpart, e.g., ARNOR takes up to $\times$1.6 the overhead of LSTM-CRF. 
DENRL, though consumes more time compared to BERT+CRF, is more efficient than other DS methods using PLMs. 

\section{Related Work}

\textbf{Joint extraction.}
Entities and relations extraction is important to construct a KB.
Traditional methods treat this problem as two separate tasks, i.e., NER and RE. 
Joint extraction detects entities and their relations using a single model which effectively integrates the information of entities and relations, and therefore achieves better results in both subtasks~\cite{zheng-etal-2017-joint}.
Among them, unified methods tag entities and relations simultaneously, e.g., \citet{zheng-etal-2017-joint} propose a novel tagging scheme that converts joint extraction to a sequence labeling problem; \citet{dai2019joint} introduce query position and sequential tagging to extract overlapping relations. 
These methods avoid producing redundant information compared to the parameter-sharing neural models~\cite{gupta2016table} and require no hand-crafted features that are used in the structured systems~\cite{ yu2019joint, ren2017cotype}.


\textbf{LLMs in open-domain IE.}
LLMs have shown significant promise in the field of open-domain Information Extraction (IE). Recent surveys~\cite{xu2023large} have highlighted the diverse prompts, paradigms, backbones, and datasets employed in this domain. These studies reveal that LLMs, through their ability to understand and generate human-like text, can effectively extract structured information from unstructured data sources across various domains~\cite{pang-etal-2023-guideline}. The adaptation of LLMs to different IE tasks, including NER, RE, and event extraction, demonstrates their versatility and effectiveness. Moreover, the integration of LLMs with prompt-based learning paradigms has further enhanced their performance by leveraging the contextual knowledge embedded within these models.

\textbf{Distantly supervision.}
Previous studies on distantly-supervised NER rely on simple tricks such as early stopping~\cite{liang2020bond} and multi-type entity labeling~\cite{shang2018learning,meng2021distantly}. 
For distantly-supervised RE, existing methods include multi-instance learning~\cite{lin2016neural} that models noise problem on a bag of instances, reinforcement learning~(RL)~\cite{nooralahzadeh2019reinforcement,hu-etal-2021-gradient}, adversarial~\cite{Chen-relation-adversarial-21,hao-etal-2021-knowing} or probabilistic learning~\cite{liu-etal-2022-ceta,li-etal-2023-uncertainty} that selects trustable instances, and pattern-based methods~\cite{ratner2016data,shang-pattern-relation-22} that directly model the DS labeling process to find noise patterns, e.g., \citet{feng2018reinforcement} propose a pattern extractor based on RL and use extracted patterns as features for RE.

\textbf{Probabilistic soft logic.}
In recent years, PSL rules have been applied to machine learning topics, including model interpretability~\cite{hu2016harnessing}, probability reasoning~\cite{dellert2020exploring}, sentiment analysis~\cite{gridach2020framework}, and temporal relation extraction~\cite{zhou2021clinical}. 
We are the first to model entity-relation dependencies by designing ontology-based PSL.

\section{Conclusions}
\label{sec:conclusions}
We propose DENRL, a noise-robust learning framework for distantly-supervised joint extraction, which consists of a transformer backbone, a new loss function and a self-adaptive learning step. 
Specifically, we use Bag-of-word regularization and logic fusion to learn important relation patterns and entity-relation dependencies. 
The regularized model is able to select trustable instances and build a versatile relation pattern set. 
A self-adaptive learning procedure then iteratively improves the model and dynamically maintains trustable pattern set to reduce both entity and relation noise. 
In the future, we aim to explore more complex patterns when configuring pattern sets. We will also evaluate our framework on other tasks such as event extraction and open information extraction.

\section*{Limitations}

We incorporate a BERT backbone into a sequence tagging scheme for distantly-supervised joint extraction. 
While our current framework is built upon BERT due to computation resource constraints, it's designed with flexibility in mind. It can be easily adapted to other transformers such as GPT-2, T5, and even LLMs like Llama2, as the only difference is the computation of the transformer's final representations, which is the very first step before our architecture designs.
Though achieving state-of-the-art performance compared to other DS methods, DENRL can be computation costly due to the position-attentive loss computed on multiple start positions. 
We further conduct an efficiency analysis in Section~\ref{sec:efficiency}, demonstrating a relatively small training overhead of DENRL compared to other DS methods using PLMs.

Moreover, we focus on relations within a sentence and regard words between an entity pair as relation patterns. 
In our future work, we aim to consider relations beyond the sentence boundary for DS joint extraction to better adapt to real-world information extraction scenarios.

Furthermore, although our OLF is a one-time effort and can benefit future training, it is still hand-crafted based on ontology, and we aim to design a probabilistic method such as model uncertainty to quantify more comprehensive underlying relation-entity dependencies in the future.

\section*{Acknowledgment}
This research was supported by the National Science Foundation under Grants CNS Career 2230968, CPS 2230969, CNS 2300525, CNS 2343653, CNS 2312397.

\bibliography{anthology,custom}

\begin{thebibliography}{53}
\expandafter\ifx\csname natexlab\endcsname\relax\def\natexlab#1{#1}\fi

\bibitem[{Bach et~al.(2017)Bach, Broecheler, Huang, and Getoor}]{bach2017hinge}
Stephen~H. Bach, Matthias Broecheler, Bert Huang, and Lise Getoor. 2017.
\newblock \href {http://jmlr.org/papers/v18/15-631.html} {Hinge-loss markov random fields and probabilistic soft logic}.
\newblock \emph{J. Mach. Learn. Res.}, 18:109:1--109:67.

\bibitem[{Chen et~al.(2020)Chen, Lan, Du, and Lobanov}]{chen2020joint}
Miao Chen, Ganhui Lan, Fang Du, and Victor Lobanov. 2020.
\newblock \href {https://doi.org/10.18653/v1/2020.clinicalnlp-1.26} {Joint learning with pre-trained transformer on named entity recognition and relation extraction tasks for clinical analytics}.
\newblock In \emph{Proceedings of the 3rd Clinical Natural Language Processing Workshop}, pages 234--242, Online. Association for Computational Linguistics.

\bibitem[{Chen et~al.(2021)Chen, Shi, Liu, Tang, Shao, Chen, and Zhuang}]{Chen-relation-adversarial-21}
Tao Chen, Haochen Shi, Liyuan Liu, Siliang Tang, Jian Shao, Zhigang Chen, and Yueting Zhuang. 2021.
\newblock \href {https://ojs.aaai.org/index.php/AAAI/article/view/17501} {Empower distantly supervised relation extraction with collaborative adversarial training}.
\newblock In \emph{{AAAI} 2021}, pages 12675--12682. {AAAI} Press.

\bibitem[{Dai et~al.(2019)Dai, Xiao, Lyu, Dou, She, and Wang}]{dai2019joint}
Dai Dai, Xinyan Xiao, Yajuan Lyu, Shan Dou, Qiaoqiao She, and Haifeng Wang. 2019.
\newblock \href {https://doi.org/10.1609/aaai.v33i01.33016300} {Joint extraction of entities and overlapping relations using position-attentive sequence labeling}.
\newblock In \emph{{AAAI} 2019}, pages 6300--6308. {AAAI} Press.

\bibitem[{Dellert(2020)}]{dellert2020exploring}
Johannes Dellert. 2020.
\newblock \href {http://arxiv.org/abs/2004.07000} {Exploring probabilistic soft logic as a framework for integrating top-down and bottom-up processing of language in a task context}.
\newblock \emph{CoRR}, abs/2004.07000.

\bibitem[{Devlin et~al.(2019)Devlin, Chang, Lee, and Toutanova}]{devlin-etal-2019-bert}
Jacob Devlin, Ming-Wei Chang, Kenton Lee, and Kristina Toutanova. 2019.
\newblock \href {https://doi.org/10.18653/v1/N19-1423} {{BERT}: Pre-training of deep bidirectional transformers for language understanding}.
\newblock In \emph{Proceedings of the 2019 Conference of the North {A}merican Chapter of the Association for Computational Linguistics: Human Language Technologies, Volume 1 (Long and Short Papers)}, pages 4171--4186, Minneapolis, Minnesota. Association for Computational Linguistics.

\bibitem[{Ellis et~al.(2013)Ellis, Getman, Mott, Li, Griffitt, Strassel, and Wright}]{KBP_slot}
Joe Ellis, Jeremy Getman, Justin Mott, Xuansong Li, Kira Griffitt, Stephanie~M. Strassel, and Jonathan Wright. 2013.
\newblock \href {https://tac.nist.gov/publications/2013/additional.papers/KBP2013\_annotation\_overview.TAC2013.proceedings.pdf} {Linguistic resources for 2013 knowledge base population evaluations}.
\newblock In \emph{Proceedings of the Sixth Text Analysis Conference, {TAC} 2013, Gaithersburg, Maryland, USA, November 18-19, 2013}. {NIST}.

\bibitem[{Feng et~al.(2018)Feng, Huang, Zhao, Yang, and Zhu}]{feng2018reinforcement}
Jun Feng, Minlie Huang, Li~Zhao, Yang Yang, and Xiaoyan Zhu. 2018.
\newblock \href {https://www.aaai.org/ocs/index.php/AAAI/AAAI18/paper/view/17151} {Reinforcement learning for relation classification from noisy data}.
\newblock In \emph{{AAAI} 2018}, pages 5779--5786. {AAAI} Press.

\bibitem[{Gridach(2020)}]{gridach2020framework}
Mourad Gridach. 2020.
\newblock \href {https://doi.org/10.1016/j.asoc.2020.106232} {A framework based on (probabilistic) soft logic and neural network for {NLP}}.
\newblock \emph{Appl. Soft Comput.}, 93:106232.

\bibitem[{Gupta et~al.(2016)Gupta, Sch{\"u}tze, and Andrassy}]{gupta2016table}
Pankaj Gupta, Hinrich Sch{\"u}tze, and Bernt Andrassy. 2016.
\newblock \href {https://aclanthology.org/C16-1239} {Table filling multi-task recurrent neural network for joint entity and relation extraction}.
\newblock In \emph{Proceedings of {COLING} 2016, the 26th International Conference on Computational Linguistics: Technical Papers}, pages 2537--2547, Osaka, Japan. The COLING 2016 Organizing Committee.

\bibitem[{Hao et~al.(2021)Hao, Yu, and Hu}]{hao-etal-2021-knowing}
Kailong Hao, Botao Yu, and Wei Hu. 2021.
\newblock \href {https://doi.org/10.18653/v1/2021.emnlp-main.761} {Knowing false negatives: An adversarial training method for distantly supervised relation extraction}.
\newblock In \emph{Proceedings of the 2021 Conference on Empirical Methods in Natural Language Processing}, pages 9661--9672, Online and Punta Cana, Dominican Republic. Association for Computational Linguistics.

\bibitem[{Hu et~al.(2021)Hu, Zhang, Yang, Li, Lin, Wen, and Yu}]{hu-etal-2021-gradient}
Xuming Hu, Chenwei Zhang, Yawen Yang, Xiaohe Li, Li~Lin, Lijie Wen, and Philip~S. Yu. 2021.
\newblock \href {https://doi.org/10.18653/v1/2021.emnlp-main.216} {Gradient imitation reinforcement learning for low resource relation extraction}.
\newblock In \emph{Proceedings of the 2021 Conference on Empirical Methods in Natural Language Processing}, pages 2737--2746, Online and Punta Cana, Dominican Republic. Association for Computational Linguistics.

\bibitem[{Hu et~al.(2016)Hu, Ma, Liu, Hovy, and Xing}]{hu2016harnessing}
Zhiting Hu, Xuezhe Ma, Zhengzhong Liu, Eduard~H. Hovy, and Eric~P. Xing. 2016.
\newblock \href {https://doi.org/10.18653/v1/p16-1228} {Harnessing deep neural networks with logic rules}.
\newblock In \emph{Proceedings of the 54th Annual Meeting of the Association for Computational Linguistics, {ACL} 2016, August 7-12, 2016, Berlin, Germany, Volume 1: Long Papers}. The Association for Computer Linguistics.

\bibitem[{Jia et~al.(2019{\natexlab{a}})Jia, Dai, Xiao, and Wu}]{jia-etal-2019-arnor}
Wei Jia, Dai Dai, Xinyan Xiao, and Hua Wu. 2019{\natexlab{a}}.
\newblock \href {https://doi.org/10.18653/v1/P19-1135} {{ARNOR}: Attention regularization based noise reduction for distant supervision relation classification}.
\newblock In \emph{Proceedings of the 57th Annual Meeting of the Association for Computational Linguistics}, pages 1399--1408, Florence, Italy. Association for Computational Linguistics.

\bibitem[{Jia et~al.(2019{\natexlab{b}})Jia, Dai, Xiao, and Wu}]{jia2019arnor}
Wei Jia, Dai Dai, Xinyan Xiao, and Hua Wu. 2019{\natexlab{b}}.
\newblock \href {https://doi.org/10.18653/v1/P19-1135} {{ARNOR}: Attention regularization based noise reduction for distant supervision relation classification}.
\newblock In \emph{Proceedings of the 57th Annual Meeting of the Association for Computational Linguistics}, pages 1399--1408, Florence, Italy. Association for Computational Linguistics.

\bibitem[{Kirsch et~al.(2020)Kirsch, Niyazova, Mock, and R{\"u}ping}]{kirsch2020noise}
Birgit Kirsch, Zamira Niyazova, Michael Mock, and Stefan R{\"u}ping. 2020.
\newblock Noise reduction in distant supervision for relation extraction using probabilistic soft logic.
\newblock In \emph{Machine Learning and Knowledge Discovery in Databases: International Workshops of ECML PKDD 2019, W{\"u}rzburg, Germany, September 16--20, 2019, Proceedings, Part II}, pages 63--78. Springer.

\bibitem[{Lafferty et~al.(2001)Lafferty, McCallum, and Pereira}]{lafferty2001conditional}
John~D. Lafferty, Andrew McCallum, and Fernando C.~N. Pereira. 2001.
\newblock Conditional random fields: Probabilistic models for segmenting and labeling sequence data.
\newblock In \emph{Proceedings of the Eighteenth International Conference on Machine Learning}, ICML '01, page 282–289, San Francisco, CA, USA. Morgan Kaufmann Publishers Inc.

\bibitem[{Li et~al.(2022)Li, Li, Ni, and McAuley}]{li-etal-2022-share}
Shuyang Li, Yufei Li, Jianmo Ni, and Julian McAuley. 2022.
\newblock \href {https://doi.org/10.18653/v1/2022.emnlp-main.761} {{SHARE}: a system for hierarchical assistive recipe editing}.
\newblock In \emph{Proceedings of the 2022 Conference on Empirical Methods in Natural Language Processing}, pages 11077--11090, Abu Dhabi, United Arab Emirates. Association for Computational Linguistics.

\bibitem[{Li et~al.(2023{\natexlab{a}})Li, Liu, Wang, Chen, Cheng, Chen, Yu, Chen, and Liu}]{li2023glad}
Yufei Li, Yanchi Liu, Haoyu Wang, Zhengzhang Chen, Wei Cheng, Yuncong Chen, Wenchao Yu, Haifeng Chen, and Cong Liu. 2023{\natexlab{a}}.
\newblock Glad: Content-aware dynamic graphs for log anomaly detection.
\newblock \emph{arXiv preprint arXiv:2309.05953}.

\bibitem[{Li et~al.(2023{\natexlab{b}})Li, Yu, Liu, Chen, and Liu}]{li-etal-2023-uncertainty}
Yufei Li, Xiao Yu, Yanchi Liu, Haifeng Chen, and Cong Liu. 2023{\natexlab{b}}.
\newblock \href {https://doi.org/10.18653/v1/2023.acl-short.116} {Uncertainty-aware bootstrap learning for joint extraction on distantly-supervised data}.
\newblock In \emph{Proceedings of the 61st Annual Meeting of the Association for Computational Linguistics (Volume 2: Short Papers)}, pages 1349--1358, Toronto, Canada. Association for Computational Linguistics.

\bibitem[{Liang et~al.(2020)Liang, Yu, Jiang, Er, Wang, Zhao, and Zhang}]{liang2020bond}
Chen Liang, Yue Yu, Haoming Jiang, Siawpeng Er, Ruijia Wang, Tuo Zhao, and Chao Zhang. 2020.
\newblock \href {https://doi.org/10.1145/3394486.3403149} {{BOND:} bert-assisted open-domain named entity recognition with distant supervision}.
\newblock In \emph{{KDD} '20: The 26th {ACM} {SIGKDD} Conference on Knowledge Discovery and Data Mining, Virtual Event, CA, USA, August 23-27, 2020}, pages 1054--1064. {ACM}.

\bibitem[{Lin et~al.(2016)Lin, Shen, Liu, Luan, and Sun}]{lin2016neural}
Yankai Lin, Shiqi Shen, Zhiyuan Liu, Huanbo Luan, and Maosong Sun. 2016.
\newblock \href {https://doi.org/10.18653/v1/P16-1200} {Neural relation extraction with selective attention over instances}.
\newblock In \emph{Proceedings of the 54th Annual Meeting of the Association for Computational Linguistics (Volume 1: Long Papers)}, pages 2124--2133, Berlin, Germany. Association for Computational Linguistics.

\bibitem[{Lin et~al.(2020)Lin, Ji, Huang, and Wu}]{lin-etal-2020-joint}
Ying Lin, Heng Ji, Fei Huang, and Lingfei Wu. 2020.
\newblock \href {https://doi.org/10.18653/v1/2020.acl-main.713} {A joint neural model for information extraction with global features}.
\newblock In \emph{Proceedings of the 58th Annual Meeting of the Association for Computational Linguistics}, pages 7999--8009, Online. Association for Computational Linguistics.

\bibitem[{Ling and Weld(2012)}]{ling2012fine}
Xiao Ling and Daniel~S. Weld. 2012.
\newblock \href {http://www.aaai.org/ocs/index.php/AAAI/AAAI12/paper/view/5152} {Fine-grained entity recognition}.
\newblock In \emph{{AAAI} 2012}. {AAAI} Press.

\bibitem[{Liu et~al.(2017)Liu, Ren, Zhu, Zhi, Gui, Ji, and Han}]{liu-etal-2017-heterogeneous}
Liyuan Liu, Xiang Ren, Qi~Zhu, Shi Zhi, Huan Gui, Heng Ji, and Jiawei Han. 2017.
\newblock \href {https://doi.org/10.18653/v1/D17-1005} {Heterogeneous supervision for relation extraction: A representation learning approach}.
\newblock In \emph{Proceedings of the 2017 Conference on Empirical Methods in Natural Language Processing}, pages 46--56, Copenhagen, Denmark. Association for Computational Linguistics.

\bibitem[{Liu et~al.(2022)Liu, Mo, Niu, and Fan}]{liu-etal-2022-ceta}
Ruri Liu, Shasha Mo, Jianwei Niu, and Shengda Fan. 2022.
\newblock \href {https://aclanthology.org/2022.coling-1.197} {{CETA}: A consensus enhanced training approach for denoising in distantly supervised relation extraction}.
\newblock In \emph{Proceedings of the 29th International Conference on Computational Linguistics}, pages 2247--2258, Gyeongju, Republic of Korea. International Committee on Computational Linguistics.

\bibitem[{Loshchilov and Hutter(2019)}]{adamW}
Ilya Loshchilov and Frank Hutter. 2019.
\newblock \href {https://openreview.net/forum?id=Bkg6RiCqY7} {Decoupled weight decay regularization}.
\newblock In \emph{7th International Conference on Learning Representations, {ICLR} 2019, New Orleans, LA, USA, May 6-9, 2019}. OpenReview.net.

\bibitem[{Luan et~al.(2018)Luan, He, Ostendorf, and Hajishirzi}]{luan-etal-2018-multi}
Yi~Luan, Luheng He, Mari Ostendorf, and Hannaneh Hajishirzi. 2018.
\newblock \href {https://doi.org/10.18653/v1/D18-1360} {Multi-task identification of entities, relations, and coreference for scientific knowledge graph construction}.
\newblock In \emph{Proceedings of the 2018 Conference on Empirical Methods in Natural Language Processing}, pages 3219--3232, Brussels, Belgium. Association for Computational Linguistics.

\bibitem[{Ma et~al.(2021)Ma, Gui, Li, Zhang, Huang, and Zhou}]{ma-etal-2021-sent}
Ruotian Ma, Tao Gui, Linyang Li, Qi~Zhang, Xuanjing Huang, and Yaqian Zhou. 2021.
\newblock \href {https://doi.org/10.18653/v1/2021.acl-long.484} {{SENT}: {S}entence-level distant relation extraction via negative training}.
\newblock In \emph{Proceedings of the 59th Annual Meeting of the Association for Computational Linguistics and the 11th International Joint Conference on Natural Language Processing (Volume 1: Long Papers)}, pages 6201--6213, Online. Association for Computational Linguistics.

\bibitem[{Meng et~al.(2021)Meng, Zhang, Huang, Wang, Zhang, Ji, and Han}]{meng2021distantly}
Yu~Meng, Yunyi Zhang, Jiaxin Huang, Xuan Wang, Yu~Zhang, Heng Ji, and Jiawei Han. 2021.
\newblock \href {https://doi.org/10.18653/v1/2021.emnlp-main.810} {Distantly-supervised named entity recognition with noise-robust learning and language model augmented self-training}.
\newblock In \emph{Proceedings of the 2021 Conference on Empirical Methods in Natural Language Processing, {EMNLP} 2021, Virtual Event / Punta Cana, Dominican Republic, 7-11 November, 2021}, pages 10367--10378. Association for Computational Linguistics.

\bibitem[{Mintz et~al.(2009)Mintz, Bills, Snow, and Jurafsky}]{mintz2009distant}
Mike Mintz, Steven Bills, Rion Snow, and Daniel Jurafsky. 2009.
\newblock \href {https://aclanthology.org/P09-1113} {Distant supervision for relation extraction without labeled data}.
\newblock In \emph{Proceedings of the Joint Conference of the 47th Annual Meeting of the {ACL} and the 4th International Joint Conference on Natural Language Processing of the {AFNLP}}, pages 1003--1011, Suntec, Singapore. Association for Computational Linguistics.

\bibitem[{Nooralahzadeh et~al.(2019)Nooralahzadeh, L{\o}nning, and {\O}vrelid}]{nooralahzadeh2019reinforcement}
Farhad Nooralahzadeh, Jan~Tore L{\o}nning, and Lilja {\O}vrelid. 2019.
\newblock \href {https://doi.org/10.18653/v1/D19-6125} {Reinforcement-based denoising of distantly supervised {NER} with partial annotation}.
\newblock In \emph{Proceedings of the 2nd Workshop on Deep Learning Approaches for Low-Resource NLP, DeepLo@EMNLP-IJCNLP 2019, Hong Kong, China, November 3, 2019}, pages 225--233. Association for Computational Linguistics.

\bibitem[{Pang et~al.(2023)Pang, Cao, Ding, and Luo}]{pang-etal-2023-guideline}
Chaoxu Pang, Yixuan Cao, Qiang Ding, and Ping Luo. 2023.
\newblock \href {https://aclanthology.org/2023.emnlp-main.950} {Guideline learning for in-context information extraction}.
\newblock In \emph{Proceedings of the 2023 Conference on Empirical Methods in Natural Language Processing}, pages 15372--15389, Singapore. Association for Computational Linguistics.

\bibitem[{Qin et~al.(2018)Qin, Xu, and Wang}]{qin-etal-2018-robust}
Pengda Qin, Weiran Xu, and William~Yang Wang. 2018.
\newblock \href {https://doi.org/10.18653/v1/P18-1199} {Robust distant supervision relation extraction via deep reinforcement learning}.
\newblock In \emph{Proceedings of the 56th Annual Meeting of the Association for Computational Linguistics (Volume 1: Long Papers)}, pages 2137--2147, Melbourne, Australia. Association for Computational Linguistics.

\bibitem[{Radford et~al.(2019)Radford, Wu, Child, Luan, Amodei, and Sutskever}]{radford2019language}
Alec Radford, Jeff Wu, Rewon Child, David Luan, Dario Amodei, and Ilya Sutskever. 2019.
\newblock Language models are unsupervised multitask learners.

\bibitem[{Raffel et~al.(2020)Raffel, Shazeer, Roberts, Lee, Narang, Matena, Zhou, Li, and Liu}]{raffel2020t5}
Colin Raffel, Noam Shazeer, Adam Roberts, Katherine Lee, Sharan Narang, Michael Matena, Yanqi Zhou, Wei Li, and Peter~J. Liu. 2020.
\newblock Exploring the limits of transfer learning with a unified text-to-text transformer.
\newblock \emph{J. Mach. Learn. Res.}, 21:140:1--140:67.

\bibitem[{Ratner et~al.(2016)Ratner, Sa, Wu, Selsam, and R{\'{e}}}]{ratner2016data}
Alexander~J. Ratner, Christopher~De Sa, Sen Wu, Daniel Selsam, and Christopher R{\'{e}}. 2016.
\newblock \href {https://proceedings.neurips.cc/paper/2016/hash/6709e8d64a5f47269ed5cea9f625f7ab-Abstract.html} {Data programming: Creating large training sets, quickly}.
\newblock In \emph{Advances in Neural Information Processing Systems 29: Annual Conference on Neural Information Processing Systems 2016, December 5-10, 2016, Barcelona, Spain}, pages 3567--3575.

\bibitem[{Ren et~al.(2017)Ren, Wu, He, Qu, Voss, Ji, Abdelzaher, and Han}]{ren2017cotype}
Xiang Ren, Zeqiu Wu, Wenqi He, Meng Qu, Clare~R. Voss, Heng Ji, Tarek~F. Abdelzaher, and Jiawei Han. 2017.
\newblock \href {https://doi.org/10.1145/3038912.3052708} {Cotype: Joint extraction of typed entities and relations with knowledge bases}.
\newblock In \emph{{WWW} 2017}, pages 1015--1024. {ACM}.

\bibitem[{Riedel et~al.(2010)Riedel, Yao, and McCallum}]{riedel2010modeling}
Sebastian Riedel, Limin Yao, and Andrew McCallum. 2010.
\newblock \href {https://doi.org/10.1007/978-3-642-15939-8\_10} {Modeling relations and their mentions without labeled text}.
\newblock In \emph{Machine Learning and Knowledge Discovery in Databases, European Conference, {ECML} {PKDD} 2010, Barcelona, Spain, September 20-24, 2010, Proceedings, Part {III}}, volume 6323 of \emph{Lecture Notes in Computer Science}, pages 148--163. Springer.

\bibitem[{Rink and Harabagiu(2010)}]{rink2010utd}
Bryan Rink and Sanda~M. Harabagiu. 2010.
\newblock \href {https://aclanthology.org/S10-1057/} {{UTD:} classifying semantic relations by combining lexical and semantic resources}.
\newblock In \emph{Proceedings of the 5th International Workshop on Semantic Evaluation, SemEval@ACL 2010, Uppsala University, Uppsala, Sweden, July 15-16, 2010}, pages 256--259. The Association for Computer Linguistics.

\bibitem[{Shaalan(2014)}]{nadeau2007survey}
Khaled Shaalan. 2014.
\newblock \href {https://doi.org/10.1162/COLI\_a\_00178} {A survey of arabic named entity recognition and classification}.
\newblock \emph{Comput. Linguistics}, 40(2):469--510.

\bibitem[{Shang et~al.(2018)Shang, Liu, Gu, Ren, Ren, and Han}]{shang2018learning}
Jingbo Shang, Liyuan Liu, Xiaotao Gu, Xiang Ren, Teng Ren, and Jiawei Han. 2018.
\newblock \href {https://doi.org/10.18653/v1/d18-1230} {Learning named entity tagger using domain-specific dictionary}.
\newblock In \emph{Proceedings of the 2018 Conference on Empirical Methods in Natural Language Processing, Brussels, Belgium, October 31 - November 4, 2018}, pages 2054--2064. Association for Computational Linguistics.

\bibitem[{Shang et~al.(2022)Shang, Huang, Sun, Wei, and Mao}]{shang-pattern-relation-22}
Yuming Shang, Heyan Huang, Xin Sun, Wei Wei, and Xian{-}Ling Mao. 2022.
\newblock \href {https://doi.org/10.1016/j.ins.2021.10.047} {A pattern-aware self-attention network for distant supervised relation extraction}.
\newblock \emph{Inf. Sci.}, 584:269--279.

\bibitem[{Tan et~al.(2018)Tan, Wang, Xie, Chen, and Shi}]{tan2018deep}
Zhixing Tan, Mingxuan Wang, Jun Xie, Yidong Chen, and Xiaodong Shi. 2018.
\newblock \href {https://dl.acm.org/doi/abs/10.5555/3504035.3504639} {Deep semantic role labeling with self-attention}.
\newblock In \emph{AAAI}, AAAI'18/IAAI'18/EAAI'18. AAAI Press.

\bibitem[{Touvron et~al.(2023)Touvron, Martin, Stone, Albert, Almahairi, Babaei, Bashlykov, Batra, Bhargava, Bhosale et~al.}]{touvron2023llama}
Hugo Touvron, Louis Martin, Kevin Stone, Peter Albert, Amjad Almahairi, Yasmine Babaei, Nikolay Bashlykov, Soumya Batra, Prajjwal Bhargava, Shruti Bhosale, et~al. 2023.
\newblock Llama 2: Open foundation and fine-tuned chat models.
\newblock \emph{arXiv preprint arXiv:2307.09288}.

\bibitem[{Vaswani et~al.(2017)Vaswani, Shazeer, Parmar, Uszkoreit, Jones, Gomez, Kaiser, and Polosukhin}]{vaswani2017attention}
Ashish Vaswani, Noam Shazeer, Niki Parmar, Jakob Uszkoreit, Llion Jones, Aidan~N. Gomez, Lukasz Kaiser, and Illia Polosukhin. 2017.
\newblock \href {https://proceedings.neurips.cc/paper/2017/hash/3f5ee243547dee91fbd053c1c4a845aa-Abstract.html} {Attention is all you need}.
\newblock In \emph{Advances in Neural Information Processing Systems 30: Annual Conference on Neural Information Processing Systems 2017, December 4-9, 2017, Long Beach, CA, {USA}}, pages 5998--6008.

\bibitem[{Walker and et~al.(2006)}]{walker2006ace}
Christopher Walker and et~al. 2006.
\newblock Ace 2005 multilingual training corpus ldc2006t06.
\newblock Web Download.

\bibitem[{Wang and Pan(2020)}]{wang2020integrating}
Wenya Wang and Sinno~Jialin Pan. 2020.
\newblock Integrating deep learning with logic fusion for information extraction.
\newblock In \emph{AAAI 2020}, volume~34, pages 9225--9232.

\bibitem[{Xu et~al.(2023)Xu, Chen, Peng, Zhang, Xu, Zhao, Wu, Zheng, and Chen}]{xu2023large}
Derong Xu, Wei Chen, Wenjun Peng, Chao Zhang, Tong Xu, Xiangyu Zhao, Xian Wu, Yefeng Zheng, and Enhong Chen. 2023.
\newblock Large language models for generative information extraction: A survey.
\newblock \emph{arXiv preprint arXiv:2312.17617}.

\bibitem[{Yu et~al.(2019)Yu, Zhang, Shu, Wang, Liu, Wang, and Li}]{yu2019joint}
Bowen Yu, Zhenyu Zhang, Xiaobo Shu, Yubin Wang, Tingwen Liu, Bin Wang, and Sujian Li. 2019.
\newblock Joint extraction of entities and relations based on a novel decomposition strategy.
\newblock \emph{arXiv preprint arXiv:1909.04273}.

\bibitem[{Zheng et~al.(2017)Zheng, Wang, Bao, Hao, Zhou, and Xu}]{zheng-etal-2017-joint}
Suncong Zheng, Feng Wang, Hongyun Bao, Yuexing Hao, Peng Zhou, and Bo~Xu. 2017.
\newblock \href {https://doi.org/10.18653/v1/P17-1113} {Joint extraction of entities and relations based on a novel tagging scheme}.
\newblock In \emph{Proceedings of the 55th Annual Meeting of the Association for Computational Linguistics (Volume 1: Long Papers)}, pages 1227--1236, Vancouver, Canada. Association for Computational Linguistics.

\bibitem[{Zhong and Chen(2021)}]{zhong-chen-2021-frustratingly}
Zexuan Zhong and Danqi Chen. 2021.
\newblock \href {https://doi.org/10.18653/v1/2021.naacl-main.5} {A frustratingly easy approach for entity and relation extraction}.
\newblock In \emph{Proceedings of the 2021 Conference of the North American Chapter of the Association for Computational Linguistics: Human Language Technologies}, pages 50--61, Online. Association for Computational Linguistics.

\bibitem[{Zhou et~al.(2021)Zhou, Yan, Han, Caufield, Chang, Sun, Ping, and Wang}]{zhou2021clinical}
Yichao Zhou, Yu~Yan, Rujun Han, J.~Harry Caufield, Kai{-}Wei Chang, Yizhou Sun, Peipei Ping, and Wei Wang. 2021.
\newblock \href {https://ojs.aaai.org/index.php/AAAI/article/view/17721} {Clinical temporal relation extraction with probabilistic soft logic regularization and global inference}.
\newblock In \emph{{AAAI} 2021}, pages 14647--14655. {AAAI} Press.

\end{thebibliography}

\clearpage

\appendix

\section{Implementation Details}
\label{sec:implementation}

\subsection{Prompt for ICL}
For the two ICL-based baselines~\cite{pang-etal-2023-guideline}, we apply the following prompt template:

\begin{itemize}[leftmargin=*]
    \item \textbf{Instruction} (depends on datasets): ``Please solve the relation extraction task. Given a context, extract all the relation triplets (head entity, relationship, tail entity), where the relationship belongs to [relation ontology]./n''

    \item \textbf{Demonstrations} (examples): ``[context 1] $\rightarrow$ [extracted triplets list 1]; [context 2] $\rightarrow$ [extracted triplets list 2] .../n''

    \item \textbf{Prefix}: ``Context:/n''
\end{itemize}

We create the prompt using the template: [Instruction] + [Demonstrations] + [Prefix] + [Input context] to the Llama-2 for ICL and compare the quality of extracted triplets.

\subsection{Training Setup}
We tune hyperparameters on the validation set via grid search.
Specifically in regularization training, we find optimal parameters $\alpha$ and $\beta$ as 1 and 0.5 for our considered datasets. 
We implement DENRL and all baselines in PyTorch, using the AdamW~\cite{adamW} optimizer with a learning rate of 5e-4, a dropout rate of 0.2, and a batch size of 8. 
For instance selection, an empirical fitness threshold is set to 0.5 with the best validation F1. 
We take a maximum of 5 new patterns in a loop for each relation type. 
In the SAL stage, we run 5 epochs in the first loop, and 1 epoch in every rest loop until the validation performance converges.
We conduct experiments and record overhead on the NVIDIA A6000 Ada server.

\section{Study on normal IE datasets}
\label{sec:IE_datasets}

\begin{table}[]
\centering
\resizebox{0.5\textwidth}{!}{
\begin{tabular}{l|ccc|ccc}
\hline
\multirow{2}{*}{\textbf{Method}} & \multicolumn{3}{c|}{\textbf{ACE05}} & \multicolumn{3}{c}{\textbf{SciERC}} \\
 & \textbf{Prec.} & \textbf{Rec.} & \textbf{F1} & \textbf{Prec.} & \textbf{Rec.} & \textbf{F1} \\
\hline
Llama-ICL & 63.57 & 68.80 & 66.08 & 42.71 & 47.89 & 45.15 \\
DENRL & 69.41 & 73.03 & 71.17 & 49.36 & 51.52 & 50.42 \\
\hline
\end{tabular}
}
\caption{Performance comparison of LLM-ICL and DENRL on ACE05 and SciERC datasets.}
\label{tab:ie_comparison}
\end{table}

We also evaluate DENRL and Llama-ICL in two popular IE datasets---ACE~\cite{walker2006ace} and SciERC~\cite{luan-etal-2018-multi}.
As shown in Table~\ref{tab:ie_comparison}, 
we observe that DENRL still achieves a significant (over 5\%) improvement in F1 over Llama-ICL, suggesting the generalizability and robustness of our methodology across diverse datasets. We will enrich the evaluation in our final version with the new results.

\section{Generalizability across Backbones}
\label{sec:backbone_generalizability}

\begin{table}[t]
    \centering
    \resizebox{0.49\textwidth}{!}{
    \begin{tabular}{cccc}
    \hline
        T5-base & BERT-large & GPT2-medium & Llama2-7B \\
    \hline
        220M & 334M  & 355M & 7B \\
    \hline
    \end{tabular}}
    \caption{Number of parameters of different backbones.}
    \label{tab:backbone_size}
\end{table}

\begin{table}[t]
    \centering
    \resizebox{0.42\textwidth}{!}{
    \begin{tabular}{c|ccc}
    \hline
        \textbf{Method} & \textbf{Prec.} & \textbf{Rec.} & \textbf{F1}  \\
    \hline 
        T5+CRF & 44.73 & \textbf{75.31} & 56.12 \\
        GPT-2+CRF & 45.11 & \underline{75.19} & 56.40 \\
        BERT+CRF & 44.98 & 74.79 & 56.18 \\
    \hline
        DENRL w/ T5 & 69.05 & 67.28 & 68.15\\
        DENRL w/ GPT-2 & \textbf{70.72} & 66.49 & \textbf{68.60} \\
        DENRL & \underline{69.37} &	67.01 & \underline{68.17} \\
    \hline
    \end{tabular}}
    \caption{Comparison of results on the NYT dataset using different backbones for both normal RE and DS RE settings.}
    \label{tab:backbone}
\end{table}

To demonstrate the generalizability of our noise reduction approach, we extend DENRL to additional two transformer backbones: T5~\cite{raffel2020t5} and GPT-2~\cite{radford2019language}.
Table~\ref{tab:backbone_size} illustrates the concrete PLMs and their number of parameters for a rather fair comparison.
As shown in Table~\ref{tab:backbone}, we can see DENRL with different backbones all achieve around 68\% F1 score on the NYT dataset, significantly outperforming existing methods.
This consistent superiority highlights that the efficacy of our method is agnostic to the backbone selection.

\subsection{OLF Rules}

We create the OLF rules based on human annotation, as in this work we focus on open-domain IE scenarios and it is straightforward to identify such logic dependencies (apparently, this method can be substituted by querying LLM for domain-specific scenarios if human expertise is not available).
Table~\ref{tab:logic_NYT} and Table~\ref{tab:logic_KBP} summarize our OLF rules for the NYT and Wiki-KBP datasets, respectively.
Our OLF can be easily built given the target dataset.
It is a one-time effort and can be reused for subsequent fine-tuning.

\begin{table*}[]
    \centering
    \resizebox{0.75\textwidth}{!}{
    \begin{tabular}{l|l}
    \hline
        \textsc{Relation} & (Head) \textsc{Entity} \\
    \hline
        /people/person/nationality & PERSON \\
        /people/deceased\_person/place\_of\_death &  PERSON \\
        /location/country/capital & LOCATION \\
        /location/location/contains & LOCATION \\
        /people/person/children & PERSON \\
        /people/person/place\_of\_birth & PERSON \\
        /people/person/place\_lived & PERSON \\
        /location/administrative\_division/country & LOCATION \\
        /location/country/administrative\_divisions & LOCATION \\
        /business/person/company & PERSON \\
        /location/neighborhood/neighborhood\_of & LOCATION \\
        /business/company/place\_founded & ORGANIZATION \\
        /business/company/founders & ORGANIZATION \\
        /sports/sports\_team/location & ORGANIZATION \\
        /sports/sports\_team\_location/teams & LOCATION \\
        /business/company\_shareholder/major\_shareholder\_of & PERSON \\
        /business/company/major\_shareholders & ORGANIZATION \\
        /people/person/ethnicity & PERSON \\
        /people/ethnicity/people & LOCATION \\
        /business/company/advisors & ORGANIZATION \\
        /people/person/religion & PERSON \\
        /people/ethnicity/geographic\_distribution & LOCATION \\
        /people/person/profession & PERSON \\
        /business/company/industry & ORGANIZATION \\
    \hline
    \end{tabular}}
    \caption{Logic rules $r: \textsc{relation} \rightarrow \textsc{entity}$ based on the NYT ontology.}
    \label{tab:logic_NYT}
\end{table*}

\begin{table*}[]
    \centering
    \resizebox{0.53\textwidth}{!}{
    \begin{tabular}{l|l}
    \hline
        \textsc{Relation} & (Head) \textsc{Entity} \\
    \hline
        per:country\_of\_birth & PERSON \\
        per:countries\_of\_residence & PERSON \\
        per:country\_of\_death & PERSON \\
        per:children & PERSON \\
        per:parents & PERSON \\
        per:religion & PERSON \\
        per:employee\_or\_member\_of & PERSON \\
        org:founded\_by & ORGANIZATION \\
        org:parents & ORGANIZATION \\
        org:shareholders & ORGANIZATION \\
        org:subsidiaries & ORGANIZATION \\
        org:member\_of & ORGANIZATION \\
    \hline
    \end{tabular}}
    \caption{Logic rules $r: \textsc{relation} \rightarrow \textsc{entity}$ based on the Wiki-KBP ontology.}
    \label{tab:logic_KBP}
\end{table*}

\section{Case Study}

\begin{table}[t]
    \centering
    \resizebox{0.47\textwidth}{!}{
    \begin{tabular}{ll}
        \hline
        \multicolumn{2}{l}{\textsc{Relation}: \textcolor{red}{\emph{Contains}} (left: $u$, right: pattern)} \\
        0.749 & $e_2$\textcolor{cyan}{\emph{, section of}} $e_1$ \\ 
        0.692 & $e_2$\textcolor{cyan}{\emph{, the capital of}} $e_1$ \\
        ... & ...  \\
        0.548 & $e_2$\textcolor{violet}{\emph{, district of}} $e_1$ \\ 
        0.554 & $e_2$ \textcolor{violet}{\emph{and other areas of}} $e_1$ \\
        0.539 & $e_2$ \textcolor{violet}{\emph{and elsewhere in the}} $e_1$ \\
        \hline
        \multicolumn{2}{l}{\textsc{Relation}: \textcolor{red}{\emph{Company\_worked}} (left: $u$, right: pattern)} \\
        0.667 & $e_1$\textcolor{cyan}{\emph{, the chief executive of}} $e_1$ \\
        0.673 & $e_2$ \textcolor{cyan}{\emph{attorney general,}} $e_1$ \\
        ... & ... \\
        0.595 & $e_1$\textcolor{violet}{\emph{, the president of the}} $e_2$ \\
        0.513 & $e_1$\textcolor{violet}{\emph{, an economist at the}} $e_2$ \\
        0.526 & $e_1$\textcolor{violet}{\emph{, the chairman and chief executive of}} $e_2$ \\
        \hline
    \end{tabular}}
    \caption{Pattern examples including \textcolor{cyan}{high-frequency} and top \textcolor{violet}{long-tail} patterns (right) and corresponding average fitness scores (left).}
    \label{tab:patterns}
\end{table}

To show that BR explores versatile patterns to enrich pattern set $\mathcal{P}$, we summarize both high-frequency patterns obtained by IDR and meaningful long-tail patterns discovered during SAL, and statistic their average fitness (see Table~\ref{tab:patterns}). 
Some long-tail patterns are not similar syntactically but still have over 0.5 average fitness scores, meaning the model learns useful semantic correlations between related feature words.



\end{document}